\documentclass[11pt]{article}

\usepackage[final]{acl}

\usepackage{times}
\usepackage{latexsym}

\usepackage[T1]{fontenc}
\usepackage[utf8]{inputenc}

\usepackage{microtype}

\usepackage{inconsolata}

\usepackage{graphicx}
\usepackage{booktabs}
\usepackage{enumitem}

\title{Cross-Lingual Stability of LLM Judges Under Controlled Generation: Evidence from Finno-Ugric Languages}

\author{
    Isaac Chung\textsuperscript{1},
    Linda Freienthal\textsuperscript{1}
    \\
    \\
    {\textsuperscript{1}Zendesk}
    \\
    {\texttt{first.last@zendesk.com}}
}

\begin{document}
\maketitle

\begin{abstract}

Cross-lingual evaluation of large language models (LLMs) typically conflates two sources of variance: genuine model performance differences and measurement instability. We investigate evaluation reliability by holding generation conditions constant while varying target language. Using synthetic customer-support dialogues generated with identical parameters across Estonian, Finnish, and Hungarian, we test whether automatic metrics and LLM-as-a-judge scoring produce stable model rankings across these morphologically rich, related Finno-Ugric languages. With a small set of Estonian native speaker annotations as a reference point, we find systematic ranking instabilities: surface-level metrics (lexical diversity, surface and semantic similarity) maintain cross-language stability, but pragmatic judgments (coherence, instruction-following) exhibit rank inversions and near-zero correlations. Because generation is controlled, these inconsistencies reflect how judge scoring behaves differently across languages rather than true model differences.

This controlled design provides a diagnostic probe: evaluation methods that fail to maintain stability under identical generation conditions signal transfer failure before deployment. Our findings suggest that zero-shot judge transfer is unreliable for discourse-level assessment in morphologically rich languages, motivating language-specific calibration against targeted human baselines. We release our controlled generation protocol, synthetic data, and evaluation framework to enable replication across language families at \url{https://github.com/isaac-chung/cross-lingual-stability-judges}.

% Evaluating large language models (LLMs) in underrepresented languages is often constrained by a lack of human supervision, leading to a reliance on automatic metrics and LLM-as-a-judge approaches transferred from English-centric settings.
% We present \textbf{UUED} (Uncovering Unreliable Evaluation Dimensions), a diagnostic framework that interrogates evaluation reliability in morphologically rich languages.
% Using controlled synthetic dialogues in Estonian, Finnish, and Hungarian, we assess cross-language ranking stability of judge-based evaluation. We find that while LLM judges are relatively stable when assessing surface-level linguistic form, pragmatic and semantic dimensions show reduced cross-language agreement under a fixed English meta-prompt, with attenuated rank correlations and occasional rank inversions for coherence and instruction-following.
% These findings suggest that measurement transfer can be fragile—evaluation reliability may degrade before model quality—motivating standardized, language-aware reporting and minimal per-language calibration checks.
\end{abstract}

\section{Introduction}

Evaluating large language models (LLMs) in morphologically rich, underrepresented languages faces a paradox: the places that most need reliable evaluation have the least human supervision. Recent benchmarks for Finno-Ugric languages like Estonian \cite{lillepalu2025estoniannativelargelanguage}, Finnish \cite{luukkonen2023fingptlargegenerativemodels}, and Hungarian \cite{yang2025openhuevalevaluatinglargelanguage} extend coverage beyond English, yet largely inherit high-resource evaluation practices—emphasizing single-turn tasks and assuming the validity of automatic or model-based scoring whose behavior in conversational settings remains poorly understood.

\begin{figure}[ht]
\centering
\includegraphics[width=0.49\textwidth]{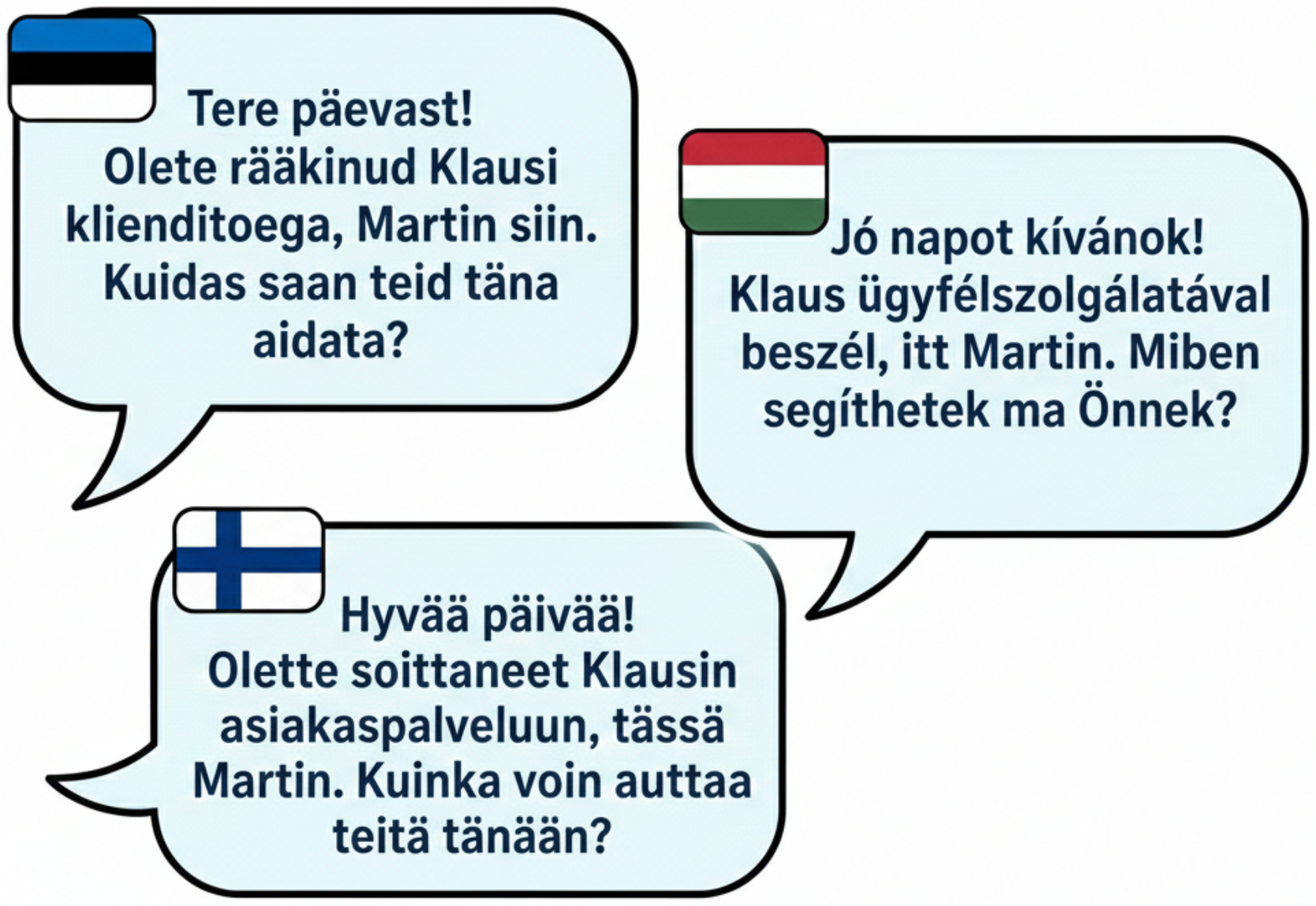}
\caption{Example opening messages in each language from the generated dialogues. In English, it reads `Good day! You have spoken to Klaus Customer Support, Martin here. How can I help you today?'.}
\label{fig:fig1}
\end{figure}

% This challenge is acute for multi-turn dialogue, where quality hinges on discourse coherence and pragmatic adequacy—dimensions that interact strongly with morphology and information structure in Finno\mbox{-}Ugric languages. Automatic metrics such as self-BLEU \cite{selfbleu2018} and BERTScore \cite{Zhang2020BERTScore} can be unreliable under rich morphology, often rewarding outputs that appear fluent but are non-idiomatic \cite{Kettunen2014ttr}. LLM-as-a-judge approaches show promise in English \cite{acl2023llmjudge}, but their cross-linguistic reliability remains largely unvalidated.

We address this validation trap through controlled  diagnostic testing: generating dialogues with identical parameters across Estonian, Finnish, and Hungarian to probe judge behavior. If rankings destabilize when only language varies, the method will fail on natural data.

Recent multilingual judge studies reveal systematic inconsistency across languages (Fleiss' $\kappa \approx$  0.3 across 25 languages; \citealt{fu2025judge}), yet the sources of this instability remain poorly understood. Our controlled generation isolates evaluation behavior from content variation to diagnose transfer failures.

Using synthetic customer-support dialogues generated with identical parameters, we first verify generation consistency through surface-level calculated metrics (lexical diversity,  surface similarity, semantic similarity), then test whether LLM-as-a-judge pragmatic assessments maintain cross-language ranking stability. 
This two-stage design isolates judge behavior: if surface properties are comparable but judge rankings diverge, the instability originates in the evaluation process rather than content variation. Our contributions are:

\begin{enumerate}[leftmargin=*,noitemsep,topsep=0pt]
    \item We demonstrate that LLM-as-a-judge coherence 
    assessment exhibits systematic rank inversions 
    ($\tau \approx 0$) across morphologically rich languages 
    under controlled generation, while surface metrics 
    maintain stability ($\tau \geq 0.76$).
    \item We provide a diagnostic methodology for detecting cross-linguistic
    ranking instabilities before large-scale deployment, 
    validated through judge ablation and prompt-language 
    sensitivity checks.
    \item We release our controlled generation protocol, synthetic dialogues, and evaluation prompts to enable replication studies in other language families. 
\end{enumerate}

\section{Methods}

\subsection{Dialogue Generation}

We generate 10K synthetic customer-support dialogues per language using parametrized templates with identical distributions across Estonian, Finnish, Hungarian, and English (40+ industries, 20+ problem types; full specifications  in \autoref{sec:appendix:convo_gen}).  
While this setup ensures semantical alignment in the prompts, we recognize that the resulting dialogue quality may vary due to the models' varying linguistic proficiencies, which may introduce subtle content variance across languages. 
English serves as a high-resource and typologically distinct anchor. By comparing Finno-Ugric outputs to this baseline, we can observe how model performance shifts when the same scenario is realized in lower-resource linguistic contexts.
Dialogues are generated end-to-end in single API calls to enable discourse-level evaluation. Code and dataset is released at \url{https://github.com/isaac-chung/cross-lingual-stability-judges}. 

\subsection{Human Annotation}

Three native Estonian speakers independently annotate 100 dialogues for coherence (conversation-level consistency) and fluency (grammatical naturalness). Inter-annotator agreement is fair to moderate ($\kappa = .385$ coherence, $\kappa = .321$ fluency), reflecting conversational evaluation subjectivity. This moderate agreement bounds expectations for automated cross-linguistic consistency—recent work shows LLM judges achieve even lower cross-language agreement \cite{fu2025judge}, highlighting the challenge of zero-shot evaluation transfer. These judgments provide a reference for interpreting automatic and judge patterns (\autoref{sec:appendix:labeling}). 

% Three native Estonian speakers independently annotate 100 randomly sampled dialogues for \textbf{logical coherence} (conversation-level consistency across turns) and \textbf{fluency} (semantic and grammatical naturalness). Inter-annotator agreement is measured using Fleiss' Kappa \cite{Fleiss1971}. 

% Inter-annotator agreement is fair to moderate ($\kappa = .385$ for coherence, $\kappa = .321$ for fluency), reflecting the inherent subjectivity of conversational evaluation. These Estonian judgments provide a reference point for interpreting automatic and judge-based scoring patterns, though the moderate agreement underscores that even native speaker evaluation contains substantial variance. Annotation guidelines are provided in \autoref{sec:appendix:labeling}.  \linda{Should we mention somewhere that we will release the golden labels?} \isaac{I feel it's implied as part of the dataset.}

% \subsection{Automatic Metrics}

% We apply reference-free automatic metrics TTR, self-BLEU \cite{selfbleu2018}, and semantic similarity to verify generation consistency across languages—a necessary precondition for interpreting cross-language evaluation stability. See \autoref{sec:appdx:auto-metrics} for full details. 

\subsection{Evaluation Framework}

We first verify generation consistency via \textit{surface-level} calculated metrics (TTR, MATTR, self-BLEU, semantic similarity; \autoref{sec:appdx:auto-metrics}), then test whether LLM-as-a-judge scoring maintains cross-language ranking stability. This two-stage design isolates judge behavior: if surface properties are comparable but judge rankings diverge, instability originates in evaluation transfer. We note, however, that this design also captures the inherent variability of generator performance across languages, allowing us to observe how the entire evaluation pipeline reacts to shifting linguistic contexts.

We use \texttt{gpt-5-mini} with default reasoning effort as an automatic judge to evaluate 100 conversations per model per language. Guided by existing works \cite{barbu2025, bae-etal-2022-building, simplestories}, the judge assigns scores for \textbf{Grammar (G)}, \textbf{Readability (R)}, \textbf{Coherence (C)}, and \textbf{Fluency (F)}. Additionally, we measure \textbf{Label Recovery Accuracy (LRA)}, which assesses instruction-following and semantic consistency by attempting to recover generation parameters from dialogue content. 
We categorize G, R, and F as \textit{surface-level} judge
metrics—evaluating grammatical correctness, lexical choice, 
and sentence-level naturalness—and C and LRA as 
\textit{pragmatic} dimensions requiring discourse-level 
reasoning about conversation flow and instruction alignment.
The judge operates zero-shot with English meta-prompts (\autoref{sec:appendix:judge}). A sensitivity check using native-language meta-prompts for Estonian showed negligible variance from English-prompt results (difference $< 0.05$; see Section \ref{sec:appendix:prompt-sensitivity} for details). An ablation across three judge models in \autoref{sec:appdx_judge_ablation} suggests that task difficulty stems from ground-truth ambiguity rather than judge capability with minimal scoring variance ($\Delta < 0.02$), supporting our choice of the cost-effective baseline model. %\linda{Did you use many judges and averaged? This paragraph is difficult for me, but if you get it, then great!} Yeah i get it - we try 3 judges, and report their own scores in Appendix G.

For each metric, we compute per-language model rankings and quantify agreement using Kendall $\tau$ (95\% bootstrap CIs, $N=1{,}500$). Rank inversions are tested via permutation. While our generation is controlled at the parameter level, observed instabilities reveal how the evaluation pipeline, comprising both the generator's output quality and the judge's scoring logic, becomes fragile when transferred to non-English contexts. %Since generation is controlled, instabilities reveal cross-linguistic ranking instabilities.

% \paragraph{Cross-language ranking stability}
% For each metric, we compute per-language per-model mean scores and ranks, then quantify cross-language agreement using Kendall $\tau$ and Spearman $\rho$ for all language pairs (et--fi, et--hu, fi--hu). We report 95\% confidence intervals via nonparametric bootstrap over dialogues ($N=1{,}500$) and count rank inversions per pair; inversion significance is assessed with a permutation test over language labels. This approach isolates evaluation-transfer effects—the `reliability gap'—by holding the domain, generator, and templates fixed. Since generation conditions are controlled, ranking instabilities reveal dimensions where the judge's internal assessment logic fails to transfer across morphologically rich languages.

\subsection{Generator Models}

We use \texttt{gpt-4.1-mini}, Llama-3.3-70B-Instruct \cite{grattafiori2024llama3herdmodels}, Mixtral-8x7B-Instruct \cite{jiang2024mixtralexperts}, Command-R, Llama-3.1-8B-Instruct \cite{grattafiori2024llama3herdmodels}, and Claude Sonnet 4, all accessed via Amazon Bedrock.\footnote{\url{https://aws.amazon.com/bedrock/}} %\linda{some references should be added perhaps?}
\section{Results}
\label{sec:results}

Our results focus on identifying systematic reliability failures in evaluation transfer rather than comparing model performance.

\begin{table*}[h!]
\centering
\resizebox{\linewidth}{!}{
\setlength{\tabcolsep}{3pt}
\begin{tabular}{l||ccc|ccc|ccc|ccc|ccc}
\hline
\textbf{Model} 
& \multicolumn{3}{c|}{\textbf{Grammar (G)}} 
& \multicolumn{3}{c|}{\textbf{Readability (R)}} 
& \multicolumn{3}{c|}{\textbf{Coherence (C)}} 
& \multicolumn{3}{c|}{\textbf{Fluency (F)}} 
& \multicolumn{3}{c}{\textbf{LRA}} \\
\hline
& et & fi & hu
& et & fi & hu
& et & fi & hu
& et & fi & hu
& et & fi & hu \\
\hline
\midrule
gpt-4.1-mini 
& $\textbf{3.17} {\scriptstyle \pm .55}$ & $\textbf{3.51} {\scriptstyle \pm .52}$ & $\textbf{3.57} {\scriptstyle \pm .50}$
& $3.63 {\scriptstyle \pm .48}$ & $\textbf{3.86} {\scriptstyle \pm .34}$ & $\textbf{3.85} {\scriptstyle \pm .36}$
& $\textbf{2.99} {\scriptstyle \pm .09}$ & $2.97 {\scriptstyle \pm .16}$ & $2.99 {\scriptstyle \pm .10}$
& $\textbf{2.35} {\scriptstyle \pm .48}$ & $\textbf{2.56} {\scriptstyle \pm .50}$ & $\textbf{2.66} {\scriptstyle \pm .48}$
& $\textbf{.62} {\scriptstyle \pm .09}$ & $.34 {\scriptstyle \pm .25}$ & $.36 {\scriptstyle \pm .26}$ \\

Llama3.3-70B-Inst. 
& $2.39 {\scriptstyle \pm .52}$ & $3.03 {\scriptstyle \pm .58}$ & $3.04 {\scriptstyle \pm .57}$
& $2.92 {\scriptstyle \pm .43}$ & $3.44 {\scriptstyle \pm .51}$ & $3.43 {\scriptstyle \pm .49}$
& $2.93 {\scriptstyle \pm .25}$ & $2.89 {\scriptstyle \pm .34}$ & $2.94 {\scriptstyle \pm .23}$
& $1.92 {\scriptstyle \pm .32}$ & $2.22 {\scriptstyle \pm .47}$ & $2.18 {\scriptstyle \pm .41}$
& $.36 {\scriptstyle \pm .20}$ & $\textbf{.62} {\scriptstyle \pm .13}$ & $.59 {\scriptstyle \pm .11}$ \\

Mixtral-8x7B-Inst. 
& $1.63 {\scriptstyle \pm .48}$ & $2.32 {\scriptstyle \pm .60}$ & $2.36 {\scriptstyle \pm .64}$
& $1.99 {\scriptstyle \pm .36}$ & $2.77 {\scriptstyle \pm .61}$ & $2.70 {\scriptstyle \pm .60}$
& $2.72 {\scriptstyle \pm .45}$ & $2.81 {\scriptstyle \pm .39}$ & $2.79 {\scriptstyle \pm .43}$
& $1.17 {\scriptstyle \pm .37}$ & $1.82 {\scriptstyle \pm .49}$ & $1.78 {\scriptstyle \pm .51}$
& $.33 {\scriptstyle \pm .24}$ & $.34 {\scriptstyle \pm .25}$ & $.33 {\scriptstyle \pm .22}$ \\

Command-R 
& $1.50 {\scriptstyle \pm .61}$ & $1.61 {\scriptstyle \pm .54}$ & $1.65 {\scriptstyle \pm .51}$
& $1.81 {\scriptstyle \pm .56}$ & $2.01 {\scriptstyle \pm .45}$ & $2.04 {\scriptstyle \pm .43}$
& $2.56 {\scriptstyle \pm .56}$ & $2.64 {\scriptstyle \pm .48}$ & $2.57 {\scriptstyle \pm .51}$
& $1.04 {\scriptstyle \pm .31}$ & $1.13 {\scriptstyle \pm .34}$ & $1.16 {\scriptstyle \pm .36}$
& $.40 {\scriptstyle \pm .20}$ & $.38 {\scriptstyle \pm .22}$ & $.34 {\scriptstyle \pm .22}$ \\

Llama3.1-8B-Inst. 
& $1.61 {\scriptstyle \pm .49}$ & $2.22 {\scriptstyle \pm .44}$ & $2.23 {\scriptstyle \pm .52}$
& $1.87 {\scriptstyle \pm .36}$ & $2.66 {\scriptstyle \pm .48}$ & $2.65 {\scriptstyle \pm .50}$
& $2.34 {\scriptstyle \pm .51}$ & $2.62 {\scriptstyle \pm .49}$ & $2.63 {\scriptstyle \pm .48}$
& $1.06 {\scriptstyle \pm .24}$ & $1.81 {\scriptstyle \pm .40}$ & $1.75 {\scriptstyle \pm .45}$
& $.30 {\scriptstyle \pm .21}$ & $.30 {\scriptstyle \pm .23}$ & $.42 {\scriptstyle \pm .20}$ \\

claude-sonnet-4 
& $3.04 {\scriptstyle \pm .50}$ & $3.18 {\scriptstyle \pm .50}$ & $3.19 {\scriptstyle \pm .47}$
& $\textbf{3.63} {\scriptstyle \pm .48}$ & $3.81 {\scriptstyle \pm .40}$ & $3.78 {\scriptstyle \pm .41}$
& $2.98 {\scriptstyle \pm .13}$ & $\textbf{2.99} {\scriptstyle \pm .09}$ & $\textbf{2.99} {\scriptstyle \pm .09}$
& $2.29 {\scriptstyle \pm .45}$ & $2.45 {\scriptstyle \pm .51}$ & $2.47 {\scriptstyle \pm .50}$
& $.45 {\scriptstyle \pm .17}$ & $.36 {\scriptstyle \pm .23}$ & $\textbf{.75} {\scriptstyle \pm .14}$ \\
\hline
\end{tabular}
}
\caption{LLM-as-a-judge evaluation of generated Estonian (et), Finnish (fi), and Hungarian (hu) dialogues. The best scores per metric and language are \textbf{bolded}.}
\label{tab:ee-llm-judge-results}
\end{table*}

\subsection{Automatic metrics reveal stable semantic content despite surface variation}

Automatic metrics (see \autoref{sec:appdx:auto-metrics} for details) reveal a nuanced picture in \autoref{tab:metrics-results}. While semantic similarity remains stable across languages (mean differences $< .03$), surface-level metrics show systematic language effects. Estonian consistently exhibits higher lexical diversity (MATTR: .48-.80) and lower repetition (Full Self-BLEU: .05-.14) compared to Finnish (MATTR: .45-.70, Self-BLEU: .11-.30) and Hungarian (MATTR: .49-.76, Self-BLEU: .22-.35) across all models. These patterns likely reflect morphological complexity differences rather than generation quality variance. 

Beyond language effects, models differ notably in lexical diversity: Llama3.1-8B shows lower MATTR (.45-.49) than Mixtral-8x7B (.70-.80). Despite these surface differences, semantic similarity remains consistent across languages.

Crucially, semantic similarity scores remain remarkably consistent (.89-.94 across all models and languages), confirming that underlying \textit{content quality} is comparable despite surface variation. This dissociation validates our experimental design for judge evaluation: generation produces semantically equivalent dialogues, but surface properties differ systematically by language.

%This validation shapes interpretation of judge-based results (\autoref{tab:ee-llm-judge-results}). Since semantic content is stable but surface properties vary, judge instabilities could stem from either: (1) surface-level bias (conflating morphological differences with quality), or (2) semantic distinctions too subtle for embedding capture. We therefore examine whether LLM-as-a-judge evaluations reflect semantic stability or surface-level bias.

\subsection{Human annotation provides a noisy reference point}

Estonian annotations yield mean scores of .842$\pm$.367 (coherence, on binary scale) and 2.108$\pm$.696 (fluency, on 0-3 scale), with fair-to-moderate agreement ($\kappa = .385$, $\kappa = .321$). Annotators report task-level coherence but reduced linguistic naturalness (\autoref{sec:appendix:labeling}). This moderate agreement bounds expectations for automated cross-linguistic consistency.

Annotators noted that dialogues were logically coherent but linguistically unnatural. Common feedback included overly formal tone, expressions that feel translated from English, and phrasing resembling 'B2 level speaker, not a native.' Frequent coherence issues included inconsistent customer names and illogical scenarios. Examples with annotator feedback are provided in \autoref{sec:appendix:labeling}.

% Human annotation of Estonian dialogues yields a mean majority-vote score of $.842 \pm .367$ for logical coherence and $2.108 \pm .696$ for fluency. These values are stable across aggregation strategies: averaging majority-vote labels per dialogue produces no meaningful difference from averaging raw annotator scores, indicating that the observed trends are not artifacts of aggregation choice. Qualitative annotator feedback reveals a consistent pattern: Dialogues are typically coherent at the task level—responses address the user's underlying intent—yet frequently exhibit reduced linguistic naturalness, including non-idiomatic constructions (see examples in \autoref{sec:appendix:labeling}). This asymmetry highlights that pragmatic adequacy is more robust than surface fluency in synthetic dialogues.

% Inter-annotator agreement is fair to moderate (Fleiss' $\kappa = .385$ for coherence, $\kappa = .321$ for fluency). This level of agreement reflects the inherent subjectivity of conversational evaluation, even among native speakers. Rather than treating these scores as a gold standard, we interpret them as a reference point for understanding evaluation variance: if human annotators show moderate disagreement, automated judges operating without language-specific calibration cannot be expected to achieve higher consistency across languages.

\subsection{LLM-as-a-judge scores diverge from human judgments and destabilize across languages}
\label{sec:results:llmjudge}

\autoref{tab:ee-llm-judge-results} shows that LLM-as-a-judge evaluations align imperfectly with human judgments in Estonian, and exhibit significant instability when extended to Finnish and Hungarian. While G and R scores remain relatively stable, scores for C, F, and LRA exhibit substantial variance across languages and models. English shows ceiling effects (C $\approx$ 2.98–3.00), limiting discriminative power but maintaining moderate ranking stability.

% \linda{"even in"? what does that hint to?}

% Critically, model rankings produced by the judge are not consistent across languages: models that score highly in Estonian are frequently ranked lower in Finnish or Hungarian, and vice versa. Because generation conditions are controlled and automatic metrics confirm comparable surface properties, these ranking inversions indicate instability in the evaluation process—a `reliability gap'—rather than genuine cross-linguistic performance differences. 

While surface metrics remain stable, coherence rankings scramble across language pairs, indicating that discourse-level assessment logic does not transfer reliably across morphologically rich languages. Label recovery accuracy (LRA) results are provided in \autoref{sec:appendix:judge}.

\subsection{Ranking stability reveals coherence breakdown}
\label{sec:results:stability}

\begin{figure*}[ht]
\centering
\includegraphics[width=\textwidth]{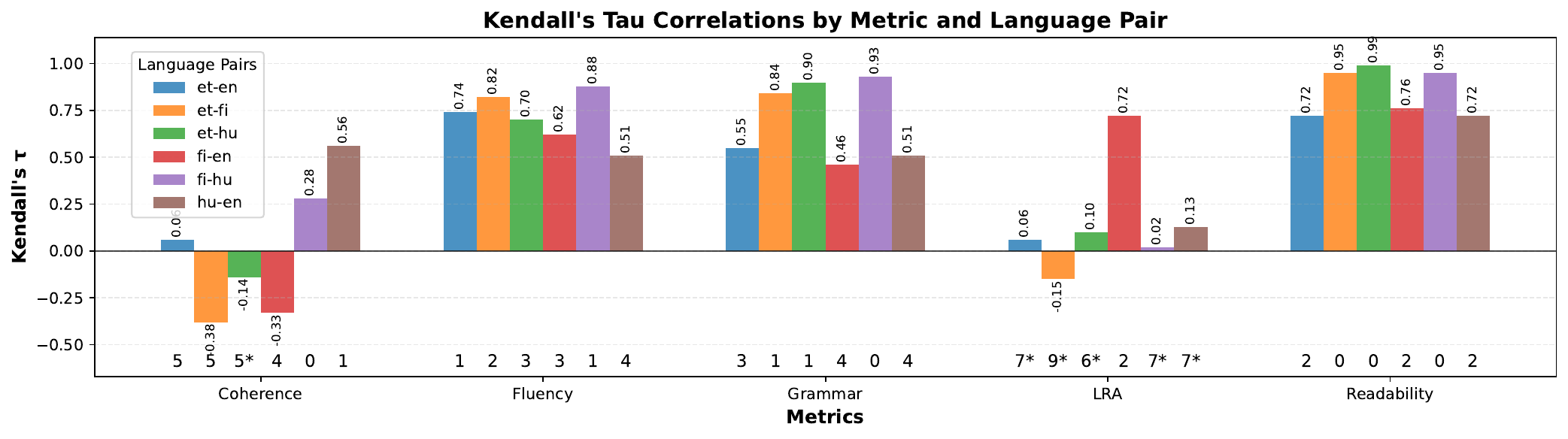}
\caption{Cross-language ranking stability measured by Kendall's $\tau$. Error bars show 95\% bootstrap confidence intervals. Numbers below bars indicate rank inversions (out of 15 possible pairwise inversions among 6 models); asterisks denote statistical significance via permutation test (* $p < 0.05$). Surface-level metrics (Grammar, Readability, Fluency) maintain high stability ($\tau \ge 0.62$) with minimal inversions. Pragmatic dimensions show systematic breakdown: Coherence exhibits near-zero or negative correlations, and LRA shows significant rank scrambling across all Finno-Ugric pairs (9*, 6*, 7* inversions). English pairs included for context, though ceiling effects limit their informativeness for Coherence.}
\label{fig:tau-correlations}
\end{figure*}

We quantify evaluation stability in \autoref{fig:tau-correlations}. The results reveal a sharp divide between surface-level and pragmatic assessment. Surface-level metrics (G, R, F) exhibit high cross-language stability ($\tau \ge .70$) with minimal rank inversions (1--3 per pair). However, Coherence shows systematic breakdown: near-zero or negative correlations across Finno-Ugric language pairs ($\tau = -.06$ for et--hu, $\tau = -.17$ for fi--hu), with significant inversions ($p = .02$) for et--hu. 
English Coherence scores show ceiling effects (mean $\approx$ 2.98–3.00), preventing meaningful ranking comparisons with English. Our analysis therefore focuses on Finno-Ugric pairs, where score variance allows for meaningful ranking comparisons.
%In contrast, Coherence maintains moderate stability for English reference pairs ($\tau \approx 0.55$), though ceiling effects in English scores (mean $\approx$ 2.98--3.00) limit discriminative power.

Since generation parameters are held constant and automatic metrics confirm comparable generation quality, these Coherence rank inversions point to judge transfer failure at the discourse level. The judge's internal discourse-level assessment logic collapses when transferred across morphologically rich languages, even among closely related language pairs. As sensitivity checks confirm that scores are robust to meta-prompt language (\autoref{sec:appendix:prompt-sensitivity}), this instability represents a fundamental breakdown in cross-linguistic evaluation reliability rather than a prompt engineering problem. These findings indicate that discourse coherence assessment—unlike surface-level grammatical or lexical evaluation—cannot be zero-shot transferred and requires language-specific calibration before deployment. Full stability analysis is provided in \autoref{app:rank-stability}.

\subsection{Meta-prompt language sensitivity}
\label{sec:appendix:prompt-sensitivity}

To ensure that the use of English-centric meta-prompts did not introduce instruction-language bias into our results, we conducted a sensitivity study on the Estonian calibration set ($N=100$). We re-evaluated the dialogues from all six generator models using a version of the LLM-as-a-judge system prompt translated into Estonian by a native speaker.

Scores produced by the native-language prompt are nearly identical to those produced by the English meta-prompt. Results suggest that the judge’s evaluation behavior is driven by its internal representation of the target language rather than the language of the instructions. Detailed results can be found in \autoref{app:meta-prompt-stability}. This rules out prompt language as the source of instability. The underlying cause remains as discussed in Section \ref{sec:results:stability}.

\begin{table*}[h!]
\centering
\resizebox{\linewidth}{!}{
\setlength{\tabcolsep}{4pt}
\begin{tabular}{l||ccc|ccc|ccc|ccc|ccc|ccc}
\hline
\textbf{Model} 
& \multicolumn{3}{c|}{\textbf{TTR}} 
& \multicolumn{3}{c|}{\textbf{MATTR}} 
& \multicolumn{3}{c|}{\textbf{Full Self-BLEU}} 
& \multicolumn{3}{c|}{\textbf{Agent Self-BLEU}} 
& \multicolumn{3}{c|}{\textbf{Client Self-BLEU}} 
& \multicolumn{3}{c}{\textbf{Intra Model Sim}} \\
\hline
& et & fi & hu
& et & fi & hu
& et & fi & hu
& et & fi & hu
& et & fi & hu
& et & fi & hu \\
\hline
gpt-4.1-mini
& $\mathbf{.80} {\scriptstyle \pm .07}$ & $.63 {\scriptstyle \pm .07}$ & $.64 {\scriptstyle \pm .07}$
& $\mathbf{.81}  {\scriptstyle \pm .06}$ & $.68 {\scriptstyle \pm .06}$ & $.71 {\scriptstyle \pm .04}$
& .10 & .19 & \textbf{.22}
& .15 & .19 & \textbf{.20}
& .12 & .11 & \textbf{.15}
& $.93 {\scriptstyle \pm .01}$ & $.92 {\scriptstyle \pm .02}$ & $\mathbf{.93} {\scriptstyle \pm .02}$ \\

Llama3.3-70B-Inst. 
& $.67 {\scriptstyle \pm .11}$ & $.52 {\scriptstyle \pm .11}$ & $.55 {\scriptstyle \pm .11}$
& $.67 {\scriptstyle \pm .11}$ & $.55 {\scriptstyle \pm .09}$ & $.59 {\scriptstyle \pm .08}$
& .14 & .30 & .35
& .26 & .34 & .36
& .25 & .18 & .25
& $\mathbf{.94} {\scriptstyle \pm .01}$ & $\mathbf{.93} {\scriptstyle \pm .02}$ & $\mathbf{.93} {\scriptstyle \pm .02}$ \\

Mixtral-8x7B-Inst. 
& $\mathbf{.80} {\scriptstyle \pm .10}$ & $\mathbf{.70} {\scriptstyle \pm .08}$ & $\mathbf{.76} {\scriptstyle \pm .08}$
& $\mathbf{.80} {\scriptstyle \pm .10}$ & $\mathbf{.70} {\scriptstyle \pm .08}$ & $\mathbf{.76} {\scriptstyle \pm .08}$
& .07 & .21 & .23
& .15 & .22 & \textbf{.20}
& .15 & .15 & .16
& $.91 {\scriptstyle \pm .02}$ & $.91 {\scriptstyle \pm .02}$ & $.90 {\scriptstyle \pm .02}$ \\

Command-R 
& $.74 {\scriptstyle \pm .10}$ & $.64 {\scriptstyle \pm .09}$ & $.60 {\scriptstyle \pm .09}$
& $.75 {\scriptstyle \pm .09}$ & $.67 {\scriptstyle \pm .07}$ & $.72 {\scriptstyle \pm .06}$
& .07 & \textbf{.11} & .31
& \textbf{.11} & \textbf{.10} & .33
& \textbf{.09} & \textbf{.07} & .25
& $.92 {\scriptstyle \pm .02}$ & $.91 {\scriptstyle \pm .02}$ & $.89 {\scriptstyle \pm .03}$ \\

Llama3.1-8B-Inst. 
& $.48 {\scriptstyle \pm .18}$ & $.42 {\scriptstyle \pm .12}$ & $.45 {\scriptstyle \pm .12}$
& $.48 {\scriptstyle \pm .19}$ & $.45 {\scriptstyle \pm .09}$ & $.49 {\scriptstyle \pm .09}$
& \textbf{.05} & .24 & .29
& .43 & .24 & .28
& .43 & .12 & .18
& $.93 {\scriptstyle \pm .01}$ & $\mathbf{.93} {\scriptstyle \pm .01}$ & $\mathbf{.93} {\scriptstyle \pm .01}$ \\

claude-sonnet-4 
& $.73 {\scriptstyle \pm .09}$ & $.59 {\scriptstyle \pm .10}$ & $.61 {\scriptstyle \pm .09}$
& $.78 {\scriptstyle \pm .06}$ & $.68 {\scriptstyle \pm .05}$ & $.72 {\scriptstyle \pm .05}$
& .12 & .19 & .26
& .19 & .19 & .25
& .12 & .13 & .19
& $.92 {\scriptstyle \pm .02}$ & $.92 {\scriptstyle \pm .02}$ & $.92 {\scriptstyle \pm .02}$ \\
\hline
\end{tabular}
}
\caption{Automatic metrics for generated Estonian (et), Finnish (fi), and Hungarian (hu) dialogues. TTR, MATTR and Intra Model Similarity show their standard deviation as well.}
\label{tab:metrics-results}
\end{table*}

\subsection{Ablation: Judge Model}
To test whether instability is specific to our chosen judge, we compared six judge models (GPT-5-mini, GPT-5.1, GPT-5.1-high, Qwen3-32B \cite{yang2025qwen3technicalreport}, Llama-4-Maverick \cite{meta2025llama4}, GPT-OSS-120B \cite{openai2025gptoss120bgptoss20bmodel}) on Finnish dialogues. All judges exhibit near-identical performance patterns with minimal variance ($\Delta < 0.02$ across categories). This suggests the instability is systematic rather than judge-specific. Full details in \autoref{sec:appdx_judge_ablation}.
\section{Discussion and Outlook}

\textbf{Surface-level evaluation transfers; discourse assessment does not.} 
Practitioners can deploy judge-based surface assessments 
(grammar, readability, fluency) for cross-linguistic 
comparison with confidence ($\tau \geq 0.70$ across 
Finno-Ugric pairs). Discourse coherence exhibits systematic 
breakdown ($\tau \approx 0$) even among related languages, 
requiring language-specific calibration.

\textbf{Controlled stability as a validity gate.} Our diagnostic 
approach provides a negative check: if an LLM 
judge produces inconsistent model rankings across languages 
under identical generation conditions, they will fare worse on natural data. 
This motivates a staged workflow: (1) verify generation 
consistency with automatic metrics, (2) collect a small 
expert sample ($N \sim 100$) in the target language, 
(3) test judge-human ranking alignment, (4) calibrate if 
correlations are weak. This prioritizes measurement 
reliability while respecting resource constraints in 
underrepresented language communities.

% UUED demonstrates that in morphologically rich and underrepresented languages, evaluation is not a neutral observer but a primary source of variance. Our results challenge the implicit assumption of cross-lingual transferability in LLM-as-a-judge frameworks. While surface-level metrics remain stable, significant rank inversions in pragmatic and semantic dimensions—specifically in LRA and Coherence—prove that a judge's internal assessment logic can collapse when transferred across related languages. 

% This `reliability gap' confirms that evaluation reliability degrades well before generative quality. Our sensitivity checks verify that this instability stems from a fundamental breakdown in cross-linguistic assessment logic rather than prompt-language bias. Consequently, LLM-as-a-judge cannot be treated as a zero-shot expert in low-resource settings; it must be systematically calibrated against human baselines. By prioritizing measurement stability over unvalidated scalability, the UUED framework ensures that multilingual benchmarking reflects genuine model capability rather than evaluation noise.

% \linda{Generated texts may seem Estonian, but it doesn't make sense. This is in appendix now}
\section*{Limitations}

%Synthetic dialogues enable controlled evaluation isolation but may exhibit stylistic homogeneity and non-idiomatic patterns not present in authentic data. Validation on natural customer support is needed to confirm reliability gaps persist in operational settings.

Synthetic dialogues enable controlled evaluation but may exhibit stylistic homogeneity and phrasing not present in real data. Validation on natural customer support scenarios is needed to confirm ranking instabilities persist in operational settings. Surface-level ranking stability suggests comparable generation quality across languages, making judge transfer failure the more likely explanation for Coherence instability. However, we cannot completely rule out discourse-level quality differences that surface metrics do not capture.

Human calibration is restricted to Estonian ($N=100$). Our 
controlled generation does not require multilingual human 
labels to detect ranking problems: if model rankings change 
when only language varies, the judge is unreliable. The 
Estonian annotations serve only to confirm that synthetic 
dialogues vary semantically and evaluate the fluency of a subset of the  synthetic 
dialogues.

We examine customer support dialogues in three related 
Finno-Ugric languages. While judge ablation 
(\autoref{sec:appdx_judge_ablation}) confirms scoring stability 
across GPT-5 variants, our findings may not hold for 
non-commercial models, other conversational domains, or 
linguistically distant languages. We focus on discourse 
coherence; other aspects like politeness conventions and 
language-specific grammatical patterns remain unexplored.

\section*{Acknowledgments}
We thank Mervi Sepp Rei, Reimo Priidik, Martin K{\"u}ngas, and Andreas Pung for labeling, Daniel Loureiro for his valuable feedback on the draft, and Joonathan M{\"a}gi, Mikk M{\"u}raus, and Kajetan Bochajczuk for their foundational work on the conversation generator. We thank Magda Kubit, Abdallah Akzouk, and Abhinay Kathuria for their support in open-source model inference. We thank the reviewers for their insightful feedback. 

\bibliography{custom}

\appendix

\newpage
\section{Automatic Metrics}
\label{sec:appdx:auto-metrics}

Here is the description of the automatic metrics.

\begin{itemize}[leftmargin=*,noitemsep,topsep=0pt]
    \item \textbf{TTR and MATTR}: we compute both simple Type--Token Ratio (TTR) (unique words / total words) and Moving Average TTR (MATTR) \cite{Kettunen2014ttr} over sliding 100-token windows for length-independent measurement of morphological variety. Higher MATTR and TTR values indicate greater lexical diversity.
    \item \textbf{Self-BLEU} \cite{selfbleu2018}: 
    Calculated at three granularity levels: full conversations, agent responses only, and client responses only—to detect formulaic patterns. We used 4-gram BLEU and NLTK's smoothing function (method4). Lower values indicate reduced repetition and greater diversity.
    \item \textbf{Intra Model Conversation Similarity} answers the question "How different are the conversations from each other?". For that we use the cosine similarity between sentence embeddings from \texttt{multilingual-e5-large-instruct} \cite{wang2024multilingual}, the highest-ranked multilingual model in MMTEB \cite{enevoldsen2025mmteb}. Lower scores indicate higher similarity (more template-like), while higher scores indicate greater conversation diversity within a model.
\end{itemize}

For calculating the values, all three languages use morphological lemmatization: EstNLTK \cite{laur-EtAl:2020:LREC} for Estonian, and Stanza \cite{qi2020stanza} for Hungarian and Finnish, with language-specific stopword filtering.

See in \autoref{tab:metrics-results} the results of the metrics per model and language.

%These metrics serve as a stability check: if they show comparable values across languages, we can attribute judge ranking instabilities to evaluation transfer rather than generation quality differences.\linda{Verify previous sentence}

%\paragraph{Interpreting Cross-Language Patterns}
%The metrics reveal systematic morphological effects across languages. Estonian's consistently higher TTR/MATTR and lower self-BLEU values likely reflect its agglutinative morphology, which naturally produces greater surface-level diversity. Finnish and Hungarian, despite also being agglutinative, show more similar patterns to each other. \linda{is it because they were calculated more similarily with stanza and with less stopwords?}Importantly, these surface differences do not indicate quality variance: semantic similarity scores remain stable (.89-.94), confirming that models generate semantically comparable content across languages. These metrics thus establish that any judge ranking instabilities cannot be attributed to systematic content quality differences, but may reflect surface-level evaluation bias. \linda{repetition}

Beyond language effects, we observe notable model differences. Llama3.1-8B shows substantially lower lexical diversity (TTR: .42-.48, MATTR: .45-.49) compared to Mixtral-8x7B-Inst. (TTR: .70-.80, MATTR: .70-.80), suggesting different training data characteristics or architectural effects on generation diversity. Command-R achieves the lowest agent-side self-BLEU scores (.10-.11 for et/fi), indicating reduced formulaic patterns in agent responses. However, all models maintain consistent semantic similarity scores across languages, confirming that surface-level differences do not translate to semantic quality variance.

\newpage
\section{Dialogue Generation} \label{sec:appendix:convo_gen}

We generate synthetic customer-support dialogues using parametrized prompt templates to create controlled test conditions across languages. Parameters control industry (40+ categories), customer problem type (20+), communication channel, agent experience, agent type, and conversation length. Crucially, \textbf{we use identical parameter distributions and generation models across all three languages}, enabling us to isolate evaluation behavior from content variation. If dialogues are generated under identical conditions but judges produce different model rankings across languages, the instability originates in the evaluation process rather than genuine performance differences.

Dialogues are generated end-to-end in a single API call, enabling evaluation of global discourse coherence rather than turn-level response quality. We generate 10K conversations per language for Estonian, Finnish, Hungarian, and English (40K total), providing sufficient scale to probe evaluation stability while remaining tractable for analysis.

\autoref{tab:convo-gen-system-prompt} and \autoref{tab:convo-gen-user-prompt} together form the dialogue generation prompt , with values of changing parameters in {curly brackets}. The parameters are sampled from the fixed sets detailed in \autoref{tab:prompt-options}. 

\begin{table*}[h]
\centering
\begin{tabular}{|p{\textwidth}|}
\hline
**Role**
You are an expert generator of customer support conversations. 
The generated conversations must stay on topic as much as possible, and mimic real life customer support interactions as much as possible.
The most important thing is for these conversations to be as realistic as possible. 

**Instructions**
These conversations are between professional agents and human customers. Customers have emotions, needs, and expectations. There are specific instructions
for each conversation that you must follow. These are for agents to follow when interacting with customers. There are also instructions
for the customer to follow.
Do you UTMOST BEST to adhere to the following instructions for generation. 
If there are more than 1 agent in the conversation, the agents turns must be sequential and must NOT interleave. For example, if agent1 and agent2 
are in the conversation, the ALLOWED turns can be:
a) agent1, customer, agent1, customer, agent2, customer, agent2; 
b) agent2, customer, agent2, customer, agent1, customer, agent1;
and the BANNED turns are:
c) agent1, customer, agent2, customer, agent1, customer, agent2;\\
\hline
\end{tabular}
\caption{System prompt for dialogue generation.}
\label{tab:convo-gen-system-prompt}
\end{table*}

\begin{table*}[h]
\centering
\begin{tabular}{|p{\textwidth}|}
\hline
(User Prompt)
Generate a chat conversation between a customer and  and \{n\_agents\} support agents. The emails of the agents are: \{agent\-emails\}.
The conversation must be in \{language\} and should be made of [\{n\_messages\}] messages.

'Klaus' is a company in the \{industry\} industry. The conversation must be tailored to the industry. For example,
use products and services that are common in the industry, and use language that is common in the industry. 
The conversation must reference at least one issue with a service, product, or policy that is relevant to the company. 

The AGENT must greet the customer. For example, using common greeting words like 'Hello' or 'Good day' in the respective language and address the customer by name, and based on the channel. 
The AGENT must use proper grammar and spelling, and must follow grammatical rules in the respective language.
The AGENT must demonstrate empathy towards the customer and must tailor the conversation to address their problems and needs.
The AGENT must use professional tone. 
\{agent\_type\}
\{problem\}
\{channel\}
\{agent\_experience\}
\\
\hline
\end{tabular}
\caption{User prompt for dialogue generation.}
\label{tab:convo-gen-user-prompt}
\end{table*}

\begin{table*}[h]
\centering
\begin{tabular}{|p{\textwidth}|}
\hline
\textbf{Industry}: manufacturing, energy production, energy management, energy technology, apparel retail, retail clothing stores, apparel manufacturing, fitness apparel retail, footwear retail, safety apparel manufacturing, home decor retail, home textiles retail, manufacturing tools, retail technology solutions, gaming technology services, transportation technology, transportation services, logistics and transportation, kitchen appliances manufacturing, utility management services, audio equipment manufacturing, e-commerce grocery retail, gambling and betting, e-commerce retail baby products, furniture retail, label manufacturing, cutlery manufacturing, bicycle manufacturing, telecommunications retail, pet retail, financial services, financial software development, gaming, retail, outdoor equipment retail, e-commerce jewelry manufacturing, retail fashion accessories, automotive parts retail, fintech services, games, e-commerce retail goods, automotive retail, coatings manufacturing, sporting goods manufacturing, e-commerce, beverage retailing, computer hardware manufacturing, automotive manufacturing, e-commerce electronics retail.\\

\textbf{Problem}: create account, delete account, edit account, switch account, check cancellation fee, delivery options, complaint, review, check invoice, get invoice, newsletter subscription, cancel order, change order, place order, check payment methods, payment issue, check refund policy, track refund, change shipping address, set up shipping address. \\

\textbf{Channel}: email, chat. \\

\textbf{Agent Experience}: junior, senior. \\

\textbf{Language}: Estonian, Finnish, Hungarian. \\

\textbf{Agent Type}: human, bot.\\

\textbf{Number of messages}: 4, 8, 12, 16.\\
\hline
\end{tabular}
\caption{Parameter options for synthetic dialogue generation. All options are sampled with equal probability, except for message length, which is weighted to favor shorter interactions $[0.4, 0.3, 0.2, 0.1]$.}
\label{tab:prompt-options}
\end{table*}

Each multi-turn conversation is generated individually in one go, similar to the method used in
PLACES \cite{chen-etal-2023-places} but without in-context learning as we operate in a data-scarce setting for underrepresented languages. Utterance-level generation strategies \cite{chen2022weakly, aher2023usinglargelanguagemodels} are not suitable for this study as we seek to evaluate full conversation generation capabilities of LLMs.
\newpage
\section{Human Labeling}
\label{sec:appendix:labeling}

\subsection{Instructions}
\label{sec:appendix:labeling-instructions}
\autoref{tab:labeler-instructions} shows detailed labeling instructions given to human labelers to evaluate the generated Estonian dialogues1. Agreement levels follow standard guidelines: $\kappa > 0.8$ (excellent), $0.6 < \kappa \leq 0.8$ (substantial), $0.4 < \kappa \leq 0.6$ (moderate), $0.2 < \kappa \leq 0.4$ (fair), and $\kappa \leq 0.2$ (poor). These expert judgments provide the calibration signal necessary to validate evaluation dimensions in morphologically rich contexts.

\begin{table*}[h]
\centering
\begin{tabular}{|p{\textwidth}|}
\hline
\textbf{Does the content make sense?} \\ 
\textbf{YES} → Questions and answers are logical, relevant to the topic. \\
\textbf{NO} → Questions and answers do not interact logically OR the issue/solution would never occur in any industry OR the agent never sends an email starting with “welcome to chat, how can I help you?” \\
\hline
\textbf{Is this fluent, human-written Estonian?} \\
\textbf{3} → Messages could pass as written by fluent speakers. \\
\textbf{2} → Majority of messages pass as written by fluent speakers, but 1–2 odd wordings and/or 1–2 grammar mistakes (e.g., wrong verb case, pronoun confusion). \\
\textbf{1} → Several odd wordings and grammar mistakes; still resembles Estonian. \\
\textbf{0} → Reading this gave me an aneurysm. \\
\hline
\end{tabular}
\caption{Detailed labeling instructions are given to human labelers for each question. }
\label{tab:labeler-instructions}
\end{table*}

\subsection{Feedback and Examples}
\label{sec:appendix:labeling-feedback}
\autoref{tab:labeler-feedback} shows one agent-customer exchange from two examples taken from Estonian dialogues that have labeler feedback. The labelers mentioned that the text contained expressions that could be used in the language but do not feel natural (e.g. gives a feeling of B2 level speaker, not a native). Many phrases felt rough or one could detect the English phrase it was translated from. This is also reflected by the fluency score, with the average of the reference label (agreement between three annotators) being 2.108±.696 on the scale of 0-3.

Regarding logical coherence, the scores are higher: The average of reference labels in conversations is .842±.367 on a binary scale. Reoccurring reasons for the negative logical coherence grade were:

\begin{itemize}[leftmargin=*,noitemsep,topsep=0pt]
    \item Inconsistent customer names or amounts of product during the conversation.
    \item The described issue is illogical. E.g., a customer bought a bicycle and now wants to know how to pay or a customer needs to return an object it has not received yet.
    \item Hallucinated words that make the entire conversation not understandable.
\end{itemize}

\begin{table*}[h!]
\centering
\begin{tabular}{|p{\textwidth}|}
\hline
\textbf{Example 1}: \\
AGENT: Tere päevast! Harald siin Klaus spordivarustuse tugitiimist. Kuidas saan teid täna aidata?\\
CUSTOMER: Tere! Tellisin hiljuti spordijalatsid, aga kahjuks pidin tellimuse tühistama. Nüüd näen, et mulle on lisatud tühistamistasu. Kas see on õigustatud?\\
\\
Fluency: 1/3\\
Coherence: 1/1\\
Feedback: Too formal. "Harold siin" is too literally translated. We usually don't say that. We say "Mina olen Harold" most likely in this context.\\
\hline
\textbf{Example 2}: \\
AGENT: Tere päevast, hea klient! Tänan, et võtsite ühendust Klaus klienditoega. Kuidas saan Teid täna aidata?
\\CUSTOMER: Tere! Ma tellisin teie poest uue mobiiltelefoni, kuid märkasin, et tarneaadress on valesti sisestatud. Kas saaksin selle muuta enne, kui tellimus välja saadetakse?
\\
Fluency: 1/3\\
Coherence: 1/1\\
Feedback: Too formal, usually these conversations are more casual. "kuid" and "ning" are usually not used in speech, only in some literature. \\
\hline
\end{tabular}
\caption{One agent-customer exchange from two example generated Estonian dialogues that have labeler feedback. Most labeler feedback flags uncommon expressions and overly formal tone, which led to lower fluency scores in those dialogues. }
\label{tab:labeler-feedback}
\end{table*}
\newpage
\section{LLM As A Judge}
\label{sec:appendix:judge}

\subsection{Instructions}
\label{sec:appendix:judge-instructions}

\autoref{tab:judge-grammar-prompt} shows the full system prompt used for the LLM-as-a-judge to evaluate the linguistic and pragmatic dimensions of the generated dialogues. This zero-shot approach uses English meta-prompts to assess performance in morphologically rich languages. As discussed in the main text, the \textbf{Label Recovery Accuracy (LRA)} dimension is further utilized as a diagnostic for instruction-following and semantic consistency by attempting to extract generation parameters from the dialogue content. The prompt used to assess LRA is shown in \autoref{tab:lra-judge-grammar-prompt}. 

\subsection{English Results}
\autoref{tab:en-llm-judge-results} shows LLM-as-a-judge results on the English dialogues. 

\begin{table*}[h!]
\centering
\resizebox{\linewidth}{!}{
\setlength{\tabcolsep}{3pt}
\begin{tabular}{l||c|c|c|c|c}
\hline
\textbf{Model} 
& \textbf{Grammar (G)} 
& \textbf{Readability (R)} 
& \textbf{Coherence (C)} 
& \textbf{Fluency (F)} 
& \textbf{LRA} \\
\hline
\midrule
claude-sonnet-4 
& $\textbf{3.99} {\scriptstyle \pm .09}$ 
& $\textbf{4.00} {\scriptstyle \pm .00}$ 
& $2.98 {\scriptstyle \pm .13}$ 
& $\textbf{3.00} {\scriptstyle \pm .00}$ 
& $\textbf{.77} {\scriptstyle \pm .16}$ \\

llama3-70b-instruct 
& $3.94 {\scriptstyle \pm .23}$ 
& $3.99 {\scriptstyle \pm .09}$ 
& $2.98 {\scriptstyle \pm .13}$ 
& $2.99 {\scriptstyle \pm .09}$ 
& $.64 {\scriptstyle \pm .10}$ \\

mixtral-8x7b-instruct 
& $3.97 {\scriptstyle \pm .18}$ 
& $3.97 {\scriptstyle \pm .18}$ 
& $2.90 {\scriptstyle \pm .29}$ 
& $2.97 {\scriptstyle \pm .16}$ 
& $.39 {\scriptstyle \pm .22}$ \\

llama3-8b-instruct 
& $3.96 {\scriptstyle \pm .20}$ 
& $3.98 {\scriptstyle \pm .13}$ 
& $2.91 {\scriptstyle \pm .29}$ 
& $2.95 {\scriptstyle \pm .22}$ 
& $.40 {\scriptstyle \pm .18}$ \\

command-r 
& $3.91 {\scriptstyle \pm .29}$ 
& $3.96 {\scriptstyle \pm .19}$ 
& $2.95 {\scriptstyle \pm .23}$ 
& $2.94 {\scriptstyle \pm .25}$ 
& $.34 {\scriptstyle \pm .22}$ \\

gpt-4.1-mini 
& $3.96 {\scriptstyle \pm .21}$ 
& $\textbf{4.00} {\scriptstyle \pm .00}$ 
& $\textbf{2.99} {\scriptstyle \pm .09}$ 
& $2.99 {\scriptstyle \pm .09}$ 
& $.37 {\scriptstyle \pm .22}$ \\
\hline
\end{tabular}
}
\caption{LLM-as-a-judge evaluation of generated English dialogues. The best scores per metric are \textbf{bolded}.}
\label{tab:en-llm-judge-results}
\end{table*}

\subsection{LRA Full Results}
\label{sec:appendix:lra}

\begin{figure*}[ht] \centering \includegraphics[width=0.8\textwidth]{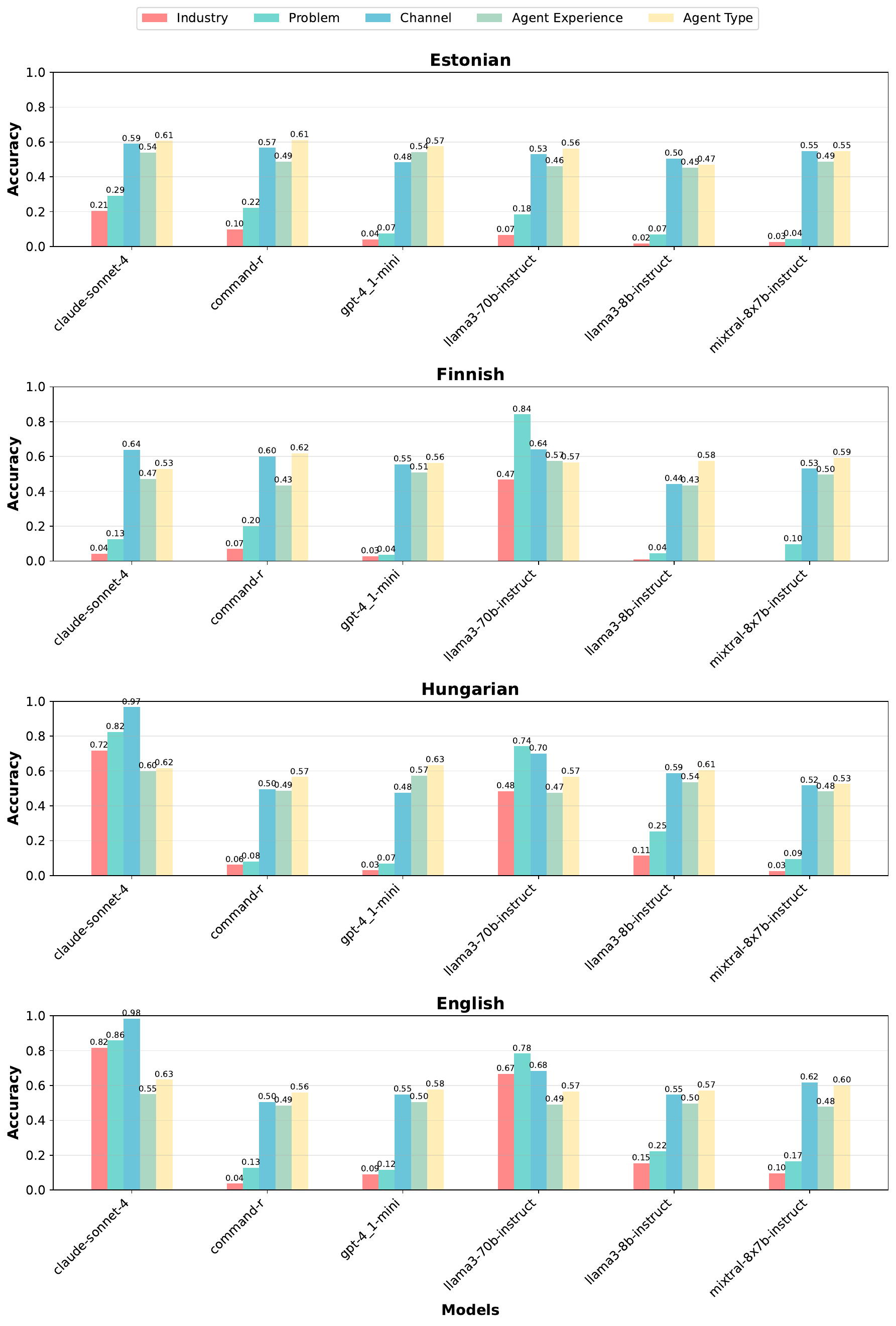} \caption{Label recovery accuracy (LRA) across categories by model for sampled dialogues in Estonian, Finnish, Hungarian, and English. Performance varies substantially by category complexity: simple binary parameters (Agent Experience, Agent Type) show consistent accuracy across all languages, while complex semantic categories (Industry: 40+ options, Problem: 20+ types) exhibit poor and inconsistent performance in all languages including English. This pattern suggests that complex parameter recovery may exceed current model capabilities regardless of target language, limiting LRA's utility as a cross-linguistic diagnostic. Unlike surface metrics and coherence assessment, where clear stability differences emerge, LRA instability appears task-dependent rather than language-dependent.} \label{fig:label_recovery_accuracy_by_category}
\end{figure*}

Label recovery accuracy measures the judge's ability to extract generation parameters from dialogue content. \autoref{fig:label_recovery_accuracy_by_category} shows performance across Estonian, Finnish, Hungarian, and English for all parameter categories.

\begin{table*}[t]
\centering
\begin{tabular}{|p{\textwidth}|}
\hline
\textbf{LLM Judge Prompt Template} \\ \hline
You are an expert linguistic judge specialising in customer-support dialogues across all languages, including Finno-Ugric languages. Your task is to rate the given dialogue according to the criteria below. Provide only the requested output format.  \\

\textbf{Grading Criteria} \\

\textbf{1. Grammaticality (G): Score 0--4}  \\
This criterion evaluates the grammatical correctness of the text, checking how free it is from grammatical errors.  \\
0: Numerous grammatical mistakes; largely unreadable.  \\
1: Significant errors that make parts difficult to understand.  \\
2: Some errors present but overall understandable.  \\
3: Minor mistakes that do not affect comprehension.  \\
4: Grammatically perfect with no mistakes.  \\

\textbf{2. Readability (R): Score 0--4}  \\
This criterion assesses ease of reading and natural flow, considering sentence length, word complexity, and overall coherence.  \\
0: Completely incoherent and unreadable.  \\
1: Very difficult to read and understand.  \\
2: Readable but requires significant effort.  \\
3: Mostly coherent, minor effort required.  \\
4: Very easy to read, natural flow.  \\

\textbf{3. Content Coherence (C): Score 0--3}  \\
3: Questions and answers are completely logical, relevant, and form a coherent dialogue flow; realistic business scenario\\
2: Questions and answers are mostly logical and relevant with minor coherence issues; plausible business scenario\\
1: Some logical connection between questions and answers but with notable coherence problems; somewhat realistic scenario\\
0: Questions and answers do not interact logically OR the issue/solution would never occur in any industry OR conversations lack proper structure\\

\textbf{4. Fluency (F): Score 0--3}  \\
How fluent and natural is this [Estonian/Finnish//English]?  \\
3: Messages could pass as written by fluent native speakers.  \\
2: Majority of messages pass as fluent, but 1--2 odd wordings and/or 1--2 grammar mistakes.  \\
1: Several odd wordings and grammar mistakes; still resembles the language.  \\
0: Extremely poor quality with pervasive errors.  \\
\hline
\end{tabular}
\caption{LLM-as-a-judge system prompt for evaluating grammar, readability, coherence, and fluency of the synthetic customer support dialogues.}
\label{tab:judge-grammar-prompt}
\end{table*}

\begin{table*}[t]
\centering
\begin{tabular}{|p{\textwidth}|}
\hline
\textbf{LLM Judge LRA Prompt Template} \\ \hline
You are an expert analyst specializing in customer support conversation classification across all languages, including Finno-Ugric languages. Your task is to classify the given conversation according to the categories below.\\

--------------------\\
CLASSIFICATION CATEGORIES\\
--------------------\\

1. Industry:  \{all possible parameters\}\\

2. Problem: Identify the primary issue or inquiry type: \{all possible parameters\}\\

3. Channel: Determine the communication method used\\
   • email: Email-based correspondence\\
   • chat: Live chat, instant messaging\\

4. Agent Experience: Assess the agent's expertise level based on responses\\
   • junior: Basic responses, may need escalation, limited problem-solving\\
   • senior: Expert responses, complex problem-solving, proactive suggestions\\

5. Agent Type: Determine if responses are from human or AI\\
   • human: Natural conversational style, empathy, contextual understanding\\
   • bot: Structured responses, consistent formatting, may lack nuance\\

Analyze the conversation carefully and provide your classification for each category along with a brief explanation.\\\\

Please classify the following customer support conversation across all required categories:\\
\{conversation\}\\\\

Provide classifications for:\\
1. Industry: Select from the specific industries listed in the system prompt (e.g., manufacturing, energy production, etc.)\\
2. Problem type: Select from the specific problem types\\
3. Channel: email or chat\\
4. Agent experience level: junior or senior\\
5. Agent type: human or bot\\

Include a brief explanation for your classification decisions.\\
\hline
\end{tabular}
\caption{LLM-as-a-judge prompt for evaluating LRA of the synthetic customer support dialogues.}
\label{tab:lra-judge-grammar-prompt}
\end{table*}
\newpage
\section{Cross-language ranking stability}
\label{app:rank-stability}
We aggregate per-language per-model means and compute rank correlations for each language pair (et--en, et--fi, et--hu, fi--en, fi--hu, hu--en). To assess whether observed order flips exceed chance, we run a permutation test (randomly reassigning language labels at the per-model level) and report 95\% bootstrap confidence intervals.
\begin{table*}[ht]
\centering
\resizebox{\linewidth}{!}{
\begin{tabular}{llccc}
\hline
\textbf{Metric} & \textbf{Pair} & \textbf{Kendall $\tau$ [95\% CI]} & \textbf{Spearman $\rho$ [95\% CI]} & \textbf{Inversions (obs, $p$)} \\
\hline
Coherence   & et--en & $0.06$ $[-0.30, 0.45]$ & $0.07$ $[-0.39, 0.52]$ & 5, $0.05$ \\
Coherence   & et--fi & $-0.38$ $[-0.65, 0.65]$ & $-0.41$ $[-0.70, 0.70]$ & 5, $0.05$ \\
Coherence   & et--hu & $-0.14$ $[-0.67, 0.29]$ & $-0.16$ $[-0.72, 0.36]$ & 5, $0.04$ \\
Coherence   & fi--en & $-0.33$ $[-0.58, -0.12]$ & $-0.38$ $[-0.65, -0.13]$ & 4, $0.10$ \\
Coherence   & fi--hu & $0.28$ $[0.00, 0.30]$ & $0.29$ $[0.00, 0.31]$ & 0, $1.00$ \\
Coherence   & hu--en & $0.56$ $[0.26, 0.60]$ & $0.64$ $[0.37, 0.78]$ & 1, $0.66$ \\
\hline
Fluency     & et--en & $0.74$ $[0.47, 1.00]$ & $0.86$ $[0.60, 1.00]$ & 1, $0.80$ \\
Fluency     & et--fi & $0.82$ $[0.60, 1.00]$ & $0.92$ $[0.77, 1.00]$ & 2, $0.44$ \\
Fluency     & et--hu & $0.70$ $[0.60, 0.87]$ & $0.83$ $[0.77, 0.94]$ & 3, $0.28$ \\
Fluency     & fi--en & $0.62$ $[0.33, 0.87]$ & $0.77$ $[0.54, 0.94]$ & 3, $0.27$ \\
Fluency     & fi--hu & $0.88$ $[0.87, 1.00]$ & $0.95$ $[0.94, 1.00]$ & 1, $0.81$ \\
Fluency     & hu--en & $0.51$ $[0.20, 0.73]$ & $0.69$ $[0.37, 0.89]$ & 4, $0.14$ \\
\hline
Grammar     & et--en & $0.55$ $[0.33, 0.73]$ & $0.73$ $[0.54, 0.83]$ & 3, $0.26$ \\
Grammar     & et--fi & $0.84$ $[0.73, 1.00]$ & $0.93$ $[0.83, 1.00]$ & 1, $0.78$ \\
Grammar     & et--hu & $0.90$ $[0.73, 1.00]$ & $0.96$ $[0.83, 1.00]$ & 1, $0.78$ \\
Grammar     & fi--en & $0.46$ $[0.20, 0.60]$ & $0.66$ $[0.31, 0.77]$ & 4, $0.13$ \\
Grammar     & fi--hu & $0.93$ $[0.73, 1.00]$ & $0.97$ $[0.89, 1.00]$ & 0, $1.00$ \\
Grammar     & hu--en & $0.51$ $[0.33, 0.60]$ & $0.71$ $[0.49, 0.77]$ & 4, $0.13$ \\
\hline
LRA         & et--en & $0.06$ $[-0.20, 0.33]$ & $0.16$ $[-0.20, 0.49]$ & 7, $0.02$ \\
LRA         & et--fi & $-0.15$ $[-0.47, 0.20]$ & $-0.17$ $[-0.60, 0.26]$ & 9, $0.01$ \\
LRA         & et--hu & $0.10$ $[-0.20, 0.33]$ & $0.07$ $[-0.31, 0.37]$ & 6, $0.03$ \\
LRA         & fi--en & $0.72$ $[0.33, 1.00]$ & $0.82$ $[0.49, 1.00]$ & 2, $0.38$ \\
LRA         & fi--hu & $0.02$ $[-0.33, 0.33]$ & $0.02$ $[-0.31, 0.49]$ & 7, $0.02$ \\
LRA         & hu--en & $0.13$ $[-0.07, 0.33]$ & $0.25$ $[-0.09, 0.54]$ & 7, $0.02$ \\
\hline
Readability & et--en & $0.72$ $[0.55, 0.83]$ & $0.84$ $[0.70, 0.93]$ & 2, $0.45$ \\
Readability & et--fi & $0.95$ $[0.87, 1.00]$ & $0.98$ $[0.94, 1.00]$ & 0, $1.00$ \\
Readability & et--hu & $0.99$ $[0.87, 1.00]$ & $1.00$ $[0.94, 1.00]$ & 0, $1.00$ \\
Readability & fi--en & $0.76$ $[0.55, 0.97]$ & $0.87$ $[0.70, 0.99]$ & 2, $0.43$ \\
Readability & fi--hu & $0.95$ $[0.87, 1.00]$ & $0.98$ $[0.94, 1.00]$ & 0, $1.00$ \\
Readability & hu--en & $0.72$ $[0.55, 0.83]$ & $0.83$ $[0.70, 0.93]$ & 2, $0.44$ \\
\hline
\end{tabular}
}
\caption{Cross-language ranking stability: Kendall $\tau$ and Spearman $\rho$ and inversion counts (obs) with permutation $p$-values. Significant inversions ($p < 0.05$) indicate that rankings are not preserved across languages under the given evaluation.}
\label{tab:rank-stability}
\end{table*}

For inversion count $n=6$ models, the maximum number of possible pairwise inversions is $n(n-1)/2=15$. An inversion count of 0 represents perfect preservation of model ranking between two languages, while 15 represents a perfect reversal. 

Across languages, surface-oriented dimensions (Grammar, Readability, Fluency) show high rank stability ($\tau$ typically $\geq$ 0.5 with non‑significant inversion counts). In contrast, pragmatic dimensions are fragile under transfer: Coherence shows attenuated or negative agreement for pairs involving Estonian (et–en/fi/hu) with marginal/significant inversion counts, while remaining stable for fi–hu. 

LRA exhibits significant inversions for several pairs, including et--en (7, $p=0.02$), et--fi (9, $p=0.01$), et--hu (6, $p=0.03$), fi--hu (7, $p=0.02$), and hu--en (7, $p=0.02$). Coherence shows marginal/significant inversions in et--en (5, $p=0.05$), et--fi (5, $p=0.05$), and et--hu (5, $p=0.04$). Because the domain and generator are held constant, instability reflects evaluation transfer rather than model content.

\section{Meta-Prompt Sensitivity}
\label{app:meta-prompt-stability}

\autoref{tab:prompt-sensitivity-results} shows that scores produced by the native-language prompt are nearly identical to those produced by the English meta-prompt. For example, the maximum variance observed for any model in any dimension remains below $0.05$.

\begin{table*}[ht]
\centering
\resizebox{\linewidth}{!}{ 
    \begin{tabular}{l|cc|cc|cc|cc}
    \hline
    \textbf{Model} & \multicolumn{2}{c|}{\textbf{Grammar (G)}} & \multicolumn{2}{c|}{\textbf{Readability (R)}} & \multicolumn{2}{c|}{\textbf{Coherence (C)}} & \multicolumn{2}{c}{\textbf{Fluency (F)}} \\
    & en & et & en & et & en & et & en & et \\
    \hline
    gpt-4.1-mini & 3.17 & 3.18 & 3.63 & 3.60 & 2.99 & 2.99 & 2.35 & 2.36 \\
    llama-3.3-70b-inst. & 2.39 & 2.38 & 2.92 & 2.98 & 2.93 & 2.94 & 1.87 & 1.86 \\
    claude-sonnet-4 & 3.04 & 3.08 & 3.63 & 3.64 & 2.98 & 2.98 & 2.29 & 2.32 \\
    mixtral-8x7b-inst. & 1.63 & 1.63 & 1.99 & 1.93 & 2.72 & 2.72 & 1.17 & 1.22 \\
    llama-3.1-8b-inst. & 1.61 & 1.64 & 1.87 & 1.88 & 2.34 & 2.24 & 1.06 & 1.09 \\
    command-r & 1.50 & 1.52 & 1.81 & 1.83 & 2.56 & 2.45 & 1.04 & 1.07 \\
    \hline 
    \end{tabular}
} 
\caption{Comparison of LLM-as-a-judge mean scores using English (en) vs. Estonian (et) meta-prompts for the Estonian calibration set. The negligible variance confirms that evaluation stability is robust to the language of instructions.}
\label{tab:prompt-sensitivity-results} 
\end{table*}

\section{Appendix: Judge Model Ablation Study}
\label{sec:appdx_judge_ablation}

We compare six judge models for Finnish conversation label recovery: GPT-5-mini (baseline), GPT-5.1 with default reasoning, GPT-5.1 with high reasoning effort, Qwen3-32B, Llama-4-Maverick, and GPT-oss-120B. Open-source LLMs are accessed via Groq\footnote{\url{https://groq.com/}}. Each judge evaluated the same six models over the sampled Finnish dialogues over the LRA categories. The same judge prompt is used from \autoref{sec:appendix:judge} across all judges.

\begin{figure*}[ht]
\centering
\includegraphics[width=\textwidth]{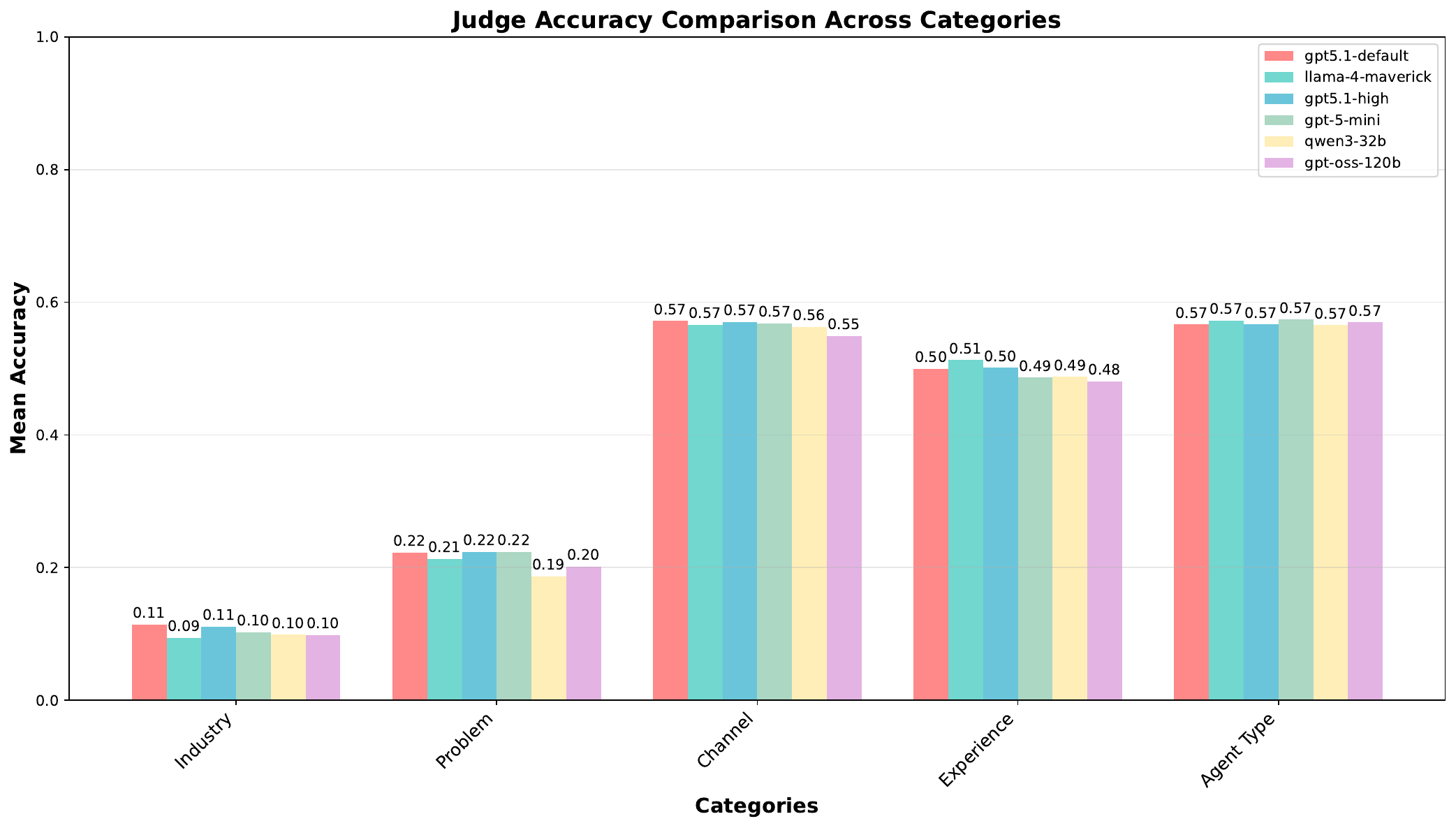}
\caption{Comparison between different LLM judges over Finnish dialogues across LLM-as-a-judge metrics.}
\label{fig:fi_judge_ablation}
\end{figure*}

\autoref{fig:fi_judge_ablation} shows accuracy by category. All six judges exhibit near-identical performance patterns, with minimal differences ($\Delta < 0.02$ across categories). Channel classification proves easiest ($\approx$55--57\%), followed by Agent Type ($\approx$57\%), Agent Experience ($\approx$48--51\%), Problem ($\approx$19--22\%), and Industry ($\approx$9--11\%).

Inter-judge agreement was assessed using Spearman correlations across all model-category pairs. Mean correlation was 0.66 across all judge comparisons, indicating moderate-to-substantial agreement while preserving meaningful judgment variance.

Three findings emerge: (1) \textbf{Model choice has minimal impact}---GPT-5.1-high performs identically to default reasoning settings, and open-source alternatives (Qwen3-32B, Llama-4-Maverick, GPT-oss-120B) achieve comparable results to proprietary models, suggesting this structured classification task does not benefit from extended reasoning or increased model scale; (2) \textbf{Trends generalize across judges}---the performance patterns observed in our main experiments with GPT-5-mini are consistently reproduced by all five alternative judges, including open-source models; (3) \textbf{Task difficulty hierarchy is judge-invariant}---all judges struggle identically with Industry/Problem categories while succeeding on Channel/Agent classifications, suggesting difficulty stems from ground-truth ambiguity rather than judge capability.

These results validate our use of GPT-5-mini as the judge throughout our main experiments, demonstrating comparable reliability to both proprietary reasoning models and open-source alternatives.

\end{document}